%% file: EuroCast2015_Kommenda.tex
\documentclass{llncs}

\usepackage{amssymb}
\usepackage{amsmath}
\usepackage{graphicx}
\usepackage{multirow}
\usepackage{subfigure}
\usepackage{multirow}
\usepackage{array}
\usepackage{siunitx}
\usepackage[table]{xcolor}

\usepackage{url}  
\begin{document}
	
\pagestyle{empty}
\mainmatter

\title{Complexity Measures for Multi-Objective Symbolic Regression}
\author{
	Michael Kommenda\inst{1,2} \and 
	Andreas Beham\inst{1,2} \and \\
	Michael Affenzeller\inst{1,2} \and
	Gabriel Kronberger\inst{1}}

\institute {
    Heuristic and Evolutionary Algorithms Laboratory\\
    School of Informatics, Communications and Media\\
    University of Applied Sciences Upper Austria\\
    Softwarepark 11, 4232 Hagenberg, Austria\\
 \vspace{0.2cm}
    \and Institute for Formal Models and Verification\\
    Johannes Kepler University  Linz\\
    Altenbergerstr. 69, 4040 Linz, Austria\\
 \vspace{0.2cm}
   \email{\{michael.kommenda, andreas.beham, michael.affenzeller,\\ gabriel.kronberger\}@fh-hagenberg.at}\\
}

\maketitle
\renewcommand{\thefootnote}{}
\footnotetext{\hspace{-0em}
	The final publication is available at \url{https://link.springer.com/chapter/10.1007/978-3-319-27340-2\_51}.
}
\renewcommand\thefootnote{\arabic{footnote}}

\begin{abstract}
	Multi-objective symbolic regression has the advantage that while the accuracy of the learned models is maximized, the complexity is automatically adapted and need not be specified a-priori. The result of the optimization is not a single solution anymore, but a whole Pareto-front describing the trade-off between accuracy and complexity.
	
	In this contribution we study which complexity measures are most appropriately used in symbolic regression when performing multi-objective optimization with NSGA-II. Furthermore, we present a novel complexity measure that includes semantic information based on the function symbols occurring in the models and test its effects on several benchmark datasets. Results comparing multiple complexity measures are presented in terms of the achieved accuracy and model length to illustrate how the search direction of the algorithm is affected.
\end{abstract}


\section{Introduction}  
\label{sec:introduction}

Symbolic regression is a data-based machine learning method, where the relation between several independent and one dependent variable is modeled. Contrary to other modeling methods the structure of the learned model is not specified a-priori, but determined during the algorithm execution. Symbolic regression problems are commonly solved by genetic programming (GP) \cite{Koza1992}, because the variable-length encoding used in GP is particularly suited for the evolution of the model structure. An expression tree encoding is frequently used in GP for symbolic regression, where every leaf node represent either a variable or a numeric constant and every internal node a mathematical function. Thus, every expression tree represents a mathematical formula that can be interpreted, validated and easily incorporated in other programs \cite{Affenzeller2013GPTP}. 

The phenomenon of bloat  in GP \cite{luke2000issues} (an increase in the average size of the individuals without an corresponding increase in fitness), overly complex and large individuals, or the excessive use of variables reduce the interpretability of symbolic regression methods. As bloat and introns are not specific to symbolic regression, but rather occur when using arbitrary-sized representations in evolutionary computation \cite{luke2002lexicographic}, several methods to limit the growth of GP individuals and to counteract bloat have been suggested previously. One approach is to specify static size and depth limits for the symbolic expression trees used in GP \cite{Poli2008} that must not be exceeded. However, these two limits are highly problem-dependent, cannot be known a-priori and must be adapted for each problem so that the trees can grow large enough to model the data accurately while unnecessary complexity is avoided.  Other methods of controlling the tree size range from dynamic size limits \cite{silva2009dynamic} or parsimony pressure methods \cite{poli2010covariant} to controlling the distribution of tree sizes \cite{dignum2008operator}.

The previously mentioned methods have been developed to limit the growth of symbolic expression trees in GP and do not include any semantic information about the mathematical formulas represented by the expression trees when performing symbolic regression. Although complexity is to some extent correlated to size, it is not necessarily the case that the complexity of a formula is reflected in its size. For example the formula $f(x)=e^{\sin{\sqrt{x}}}$ consists of three operations, one variable, and one constant, while $f(x) = 7x^2+3x+5$ consists of five operations, two variable, and three constants and is intuitively less complex.

A different approach for managing complexity and model size in symbolic regression is to use multi-objective optimization, where while maximizing the prediction accuracy the complexity is minimized \cite{smits2005pareto,luke2002lexicographic}. Hence, no size limits or other complexity related parameters must be configured and the optimization algorithm is expected to automatically evolve solutions of appropriate length and complexity. In this contribution we compare the effects on algorithm performance of several complexity and size related quality criteria for multi-objective symbolic regression on benchmark problems. 

\section{Multi-objective Symbolic Regression}
\label{sec:methods}
The nondominant sorting genetic algorithm II (NSGA-II) \cite{deb2002fast} is one of the most prominent algorithms for multi-objective optimization. It uses a novel selection mechanism based on the nondomination rank and crowding distance for selection to build a uniformly spread Pareto-optimal front. In the case of multi-objective symbolic regression the objectives to be optimized are the prediction accuracy and the complexity of the learned models. The prediction accuracy can be expressed by any error or correlation measure such as the coefficient of determination $R^2$, the mean squared error, or the mean absolute percentage error between the estimated and observed values. 

The complexity of a symbolic regression model can be calculated as the \emph{tree length} or the \emph{expressional complexity} (visitation length) \cite{smits2005pareto,keijzer2007crossover}. More sophisticated measures that also include semantics of the evolved models range from the number of included variables, or the \emph{order of nonlinearity} \cite{vladislavleva2009order} to the \emph{functional complexity} \cite{vanneschi2010measuring}. While the \emph{order of nonlinearity} and the \emph{functional complexity} express the complexity of a symbolic regression model rather accurately, they are computationally expensive to calculate. On the other hand, measures such as the \emph{tree length} or the \emph{expressional complexity} are efficiently calculated, but do not include any information except the shape of the symbolic expression tree encoding the model. 

We propose a new complexity measure that is easy to calculate and includes semantics about the regression model, so that the search direction of the multi-objective algorithm is altered towards simple and parsimonious models. The measure is calculated by recursive iteration over the symbolic expression tree and accumulates the individual complexity values for each subtree while taking into account different complexity values for the encountered function symbols. The mathematical definition for the calculation of this new complexity measure is given in Equation \ref{equ:Complexity}, where $n$ denotes a tree node and $c$ a direct child node of $n$. The complexity of the whole symbolic expression tree encoding the regression model can then be calculated by recursive application of Equation \ref{equ:Complexity} starting at the root node.

\begin{equation}
\label{equ:Complexity}
\text{Complexity}(n)  =  
	\begin{cases}
		1 & \mbox{if } n \equiv \text{constant} \\ 
		2 & \mbox{if } n \equiv \text{variable} \\ 
		\sum  \text{Complexity}(c)  & \mbox{if } n \in (+,-)\\ 
		\prod \text{Complexity}(c)+1 & \mbox{if } n \in (*,/)\\ 
		\text{Complexity}(c)^2 & \mbox{if } n \equiv \text{square}\\ 
		\text{Complexity}(c)^3 & \mbox{if } n \equiv \text{squareroot}\\ 
		2^{\text{Complexity}(c)} & \mbox{if } n \in (\sin,\cos,\tan,\exp,\log)\\ 
	\end{cases} \\
\end{equation}
\\

\begin{figure}[thbp]
	\centering
	\includegraphics[width=0.5\textwidth,clip=true]{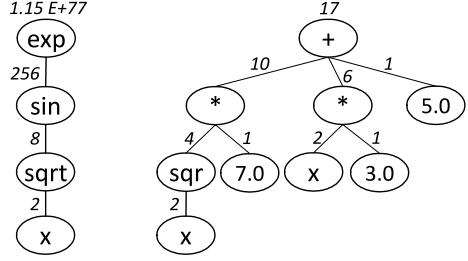}
	\caption{Symbolic expression tree representation of $f_1(x)=e^{\sin{\sqrt{x}}}$ and $f_2(x) = 7x^2+3x+5$, where the calculation steps for the complexity measure are indicated next to the arcs.}
	\label{fig:ComplexityCalculation}
\end{figure}

In the following we explore the complexity differences of two exemplary models, $f_1(x)=e^{\sin{\sqrt{x}}}$ and $f_2(x) = 7x^2+3x+5$. The corresponding symbolic expression trees representing these formulas are illustrated in Figure \ref{fig:ComplexityCalculation}. The tree length for $f_1(x)$ is $4$ and $9$ for $f_2(x)$, which indicates that representation of $f_1(x)$ is more compact. However, Figure \ref{fig:ComplexityCalculation} also shows the calculation steps for the new complexity measure according to Equation \ref{equ:Complexity}, which is iteratively applied starting at the leaf nodes. The complexity measure results in $65536$ for $f_1(x)$ and $17$ for $f_2(x)$, which reflects our intuition that $f_2(x)$ is less complex and easier to interpret than $f_1(x)$, whereas according to the tree length the contrary is true.

\section{Experiments}
\label{sec:experiments}
We used an NSGA-II algorithm to test the effects of different complexity measures for multi-objective symbolic regression. The first objective for the algorithm is the Pearson's $R^2$ correlation describing the model accuracy. Varying complexity measures such as the number of used variables (\emph{Variables}), the model length (\emph{Tree Length}), the expressional complexity (\emph{Visitation Length}) and the new complexity measure (\emph{Complexity}) described in Equation \ref{equ:Complexity} have been used as second optimization objective. Despite maximizing the Pearson's $R^2$, the results regarding accuracy are presented in terms of the normalized mean squared error (NMSE), so that for both measures smaller values are better.

NSGA-II was configured to evolve models with a maximum tree length of $100$ that are allowed to include arithmetic ($+,-,*,/$), trigonometric ($\sin$, $\cos$, $\tan$), power($^2$, $\sqrt{ }$) and exponential ($\exp, \log$) symbols and stops when the termination criterion of $200,000$ model evaluations has been reached. The benchmark problems used for testing have been selected from \cite{White:2013:GPEM,friedman1991multivariate} and the generating formulas are listed in Table \ref{tab:problems}. 

\begin{table}
	\centering
	\caption{Definition of benchmark problems.}
	\begin{tabular}{p{2.2cm}|p{0.05cm}l}
		Name 			&& Function \\ 
		\hline  && \\ [-2ex]	
		Keijzer-5 		&& $f(x_1, x_2, x_3) = 30x_1x_3/ [(x_1-10)x_2^2]$  \\ [0.5ex]
		Vladislavleva-1	&& $f(x_1, x_2)= e^{-(x_1-1)^2)} / [1.2+ (x_2-2.5)^2]$ \\[0.5ex]
		Vladislavleva-2	&& $f(x_1)=e^{-x_1} x_1^3 \cos(x) \sin(x) (\cos(x)\sin^2(x)-1)$ \\ [0.5ex]
		Vladislavleva-7	&& $f(x_1, x_2)= (x_1 - 3)(x_2 - 3) + 2 \sin((x_1 - 4)(x_2 - 4))$ \\ [0.5ex]
		Pagie-1 		&& $f(x_1,x_2)=1/[1+x_1^{-4}]+1/[1+x_2^{-4}]$ \\ [0.5ex]
		Poly-10 		&& $f(x_1 - x_{10})= x_1x_2+x_3x_4+x_5x_6+x_1x_7x_9+x_3x_6x_{10}$ \\ [0.5ex]
		Friedman-1 		&& $f(x_1 - x_{10})= 0.1 e^{4x_1} + 4/[1+e^{-20x_2}+10] + 3x_3 + 2x_2 + x_5 + \text{N}$ \\ [0.5ex]
		Friedman-2 		&& $f(x_1 - x_{10})= 10\sin(\pi x_1x_2)+20(x_3-0.5)^2+10x_4+5x_5 + \text{N}$  \\ [0.5ex]
		Tower			&& Real world data  \\  
	\end{tabular} 
	\label{tab:problems}
\end{table}

\begin{figure}[thbp]
	\centering
	\includegraphics[width=0.55\textwidth,clip=true]{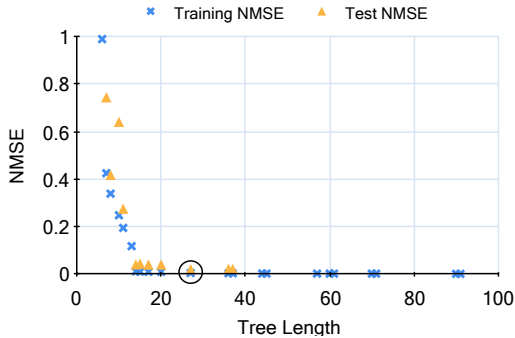}
	\caption{Exemplary Pareto-front evolved by NSGA-II showing the trade-off between accuracy in terms of the NMSE and the model length. Every model contributes to two data points, one for training and one for test evaluation (if no test evaluation is shown, the NMSE exceeds $1.0$). The \emph{best} model, which is rather small but still accurate, is encircled.}
	\label{fig:ParetoFront}
\end{figure}

When switching from single to multi-objective algorithms the result of an algorithm execution is not a single solution any more, but rather a Pareto-front showing the trade-off between accuracy and complexity of the models. An example of a Pareto-front generated by the NSGA-II is shown in Figure \ref{fig:ParetoFront}, where besides the training qualities the models' accuracies on the test set  are shown to evaluate their generalization capabilities. However, whole Pareto-fronts are hard to compare, especially in the case of symbolic regression where a high training quality doesn't necessary indicate a good model due to overfitting reasons (for example the largest models in Figure \ref{fig:ParetoFront} whose test NMSEs exceeds $1.0$ and are not displayed at all). Hence, we used only single models of a Pareto-front for algorithm comparison and used the model with the highest training accuracy. A better way for model selection would be, if an additional validation partition is defined, to use the model with the highest accuracy on this partition. 

\section{Results}
\label{sec:results}
We have performed $50$ repetitions of each NSGA-II configuration to account for the stochastic nature of the algorithm. We extracted the most accurate models, at the same time the most complex ones, from each generated Pareto-front and compared them against each other. In Table \ref{tab:results_training} the average and standard deviation of those models are displayed. Although for most of the problems the results are quite similar, there are some differences. The \emph{Complexity} configuration obtains by far the best results with the smallest variation on the Poly-10 problem. The \emph{Variables} configuration works best on the Friedman-2 problem and both configuration perfrom well on the Pagie-1 problem.
\input{results_training}

The generalization capabilities of the best models have been evaluated on a separate test partition and those results are shown in Table \ref{tab:results_test}. Every configuration produces, at least on some problems, overfit models, which is indicate by an high average normalized squared error and high standard deviations. The \emph{Tree Length} and \emph{Visitation Length} perform equally well with the exception of the Vladislavleva-1 and Pagie-1 problem. It still holds that the \emph{Complexity} and \emph{Variables} algorithm runs produced the best models on the Poly-10 and Friedman-2 problem respectively. An interesting observation is that excluding Vladislavleva-1 and Vladislavleva-2 the \emph{Complexity} measure as second objective performs either better or equally well as the other configurations. A reason for this might be that those two problems consist of complicated formulas that have to be discovered and the algorithm is not able to build that complicated yet accurate formulas. The other complexity measure that do not include semantic information about the models, have no such limitations as long as the models are compact enough. 
\input{results_test}

Next to the accuracy of the models their interpretability is of importance, because model interpretability is one of the major reasons to use symbolic regression. The lengths of the models generated by NSGA-II with varying complexity measures have been compared and the results are stated in Table \ref{tab:results_length}. It is clear that the \emph{Variables} configuration, which just counts the variable occurrences, generates by far the largest models as no selection pressure towards more parsimonious ones is applied. The \emph{Tree Length} and \emph{Visitation Length} produce the smallest models and no significant differences could be found between these two variations. Although no explicit parsimony pressure is applied to models created by NSGA-II with the new complexity measure, these are smaller than the ones produced by \emph{Variables}, but still larger than those using explicitly the tree and visitation length for optimization.  
\input{results_length}

\section{Conclusion}
\label{sec:conclusion}
In this publication we have investigated the effects of different complexity measures on multi-objective symbolic regression. Furthermore, we have presented a novel complexity measure based on the mathematical symbols occurring in the formulas. The differences with respect to the accuracies of the models on the tested benchmark problems were in most cases not significant. An exception is the new complexity measure that performs best on problems with simpler data generating formulas, but on the other hand fails to evolve well-fitting models on the most complex problems. 

When comparing the length of the evolved models to give an indication of their interpretability, the algorithm configurations explicitly using the tree length as an optimization objective generated the most parsimonious models. As previously argued and used as motivation for the development of the new complexity measure, the length is only to some extend correlated to simplicity and interpretability and more research to illustrate the differences of the complexity measures has to be performed. 

\clearpage
\subsubsection*{Acknowledgments}
The work described in this paper was done within the COMET Project Heuristic Optimization in Production and Logistics (HOPL), \#843532 funded by the Austrian Research Promotion Agency (FFG).

\bibliographystyle{splncs03}
\bibliography{eurocast2015}

\end{document}

%% file: results_training.tex
\newcolumntype{x}[1]{>{\centering\arraybackslash\hspace{0pt}}p{#1}}

\begin{table}
	\centering
	\caption{Average and standard deviation ($\mu \pm \sigma$) of the training qualities (NMSE) of the best individual for 50 repetitions of NSGA-II with varying complexity measures.}
	\begin{tabular}{p{2.3cm}|x{2.3cm}|x{2.2cm}|x{2.6cm}|x{2.2cm}	}
		Problem 		 &   Variables   		 &  Tree Length 		 & Visitation Length 	 &  Complexity \\ \hline
		&&&& \\ [-2ex]
		Keijzer-5		 & 	$0.000 \pm 0.000$	 & 	$0.003 \pm 0.003$	 & 	$0.002 \pm 0.003$	 & $0.000 \pm 0.000$ \\ 
		Vladislavleva-1	 & 	$0.002 \pm 0.002$	 & 	$0.005 \pm 0.011$	 & 	$0.004 \pm 0.004$	 & $0.002 \pm 0.002$ \\ 
		Vladislavleva-2	 & 	$0.022 \pm 0.020$	 & 	$0.018 \pm 0.016$	 & 	$0.014 \pm 0.012$	 & $0.016 \pm 0.013$ \\ 
		Vladislavleva-7	 & 	$0.111 \pm 0.020$	 & 	$0.147 \pm 0.077$	 & 	$0.116 \pm 0.027$	 & $0.115 \pm 0.027$ \\ 
		Pagie-1	 		 & 	$0.001 \pm 0.003$	 & 	$0.011 \pm 0.017$	 & 	$0.012 \pm 0.018$	 & $0.007 \pm 0.003$ \\ 
		Poly-10	 		 & 	$0.294 \pm 0.126$	 & 	$0.356 \pm 0.165$	 & 	$0.341 \pm 0.168$	 & $0.202 \pm 0.079$ \\ 
		Friedman-1	 	 & 	$0.140 \pm 0.004$	 & 	$0.157 \pm 0.019$	 & 	$0.163 \pm 0.023$	 & $0.175 \pm 0.021$ \\ 
		Friedman-2		 & 	$0.081 \pm 0.031$	 & 	$0.134 \pm 0.050$	 & 	$0.146 \pm 0.038$	 & $0.143 \pm 0.049$ \\ 
		Tower	 		 & 	$0.148 \pm 0.025$	 & 	$0.148 \pm 0.020$	 & 	$0.146 \pm 0.016$	 & $0.140 \pm 0.018$ \\ 
	\end{tabular} 
	\label{tab:results_training}
\end{table}

%% file: results_test.tex
\begin{table}
	\centering
	\caption{Average and standard deviation ($\mu \pm \sigma$) of the quality of the best training individual evaluated on the test partition for 50 repetitions of the multi-objective symbolic regression algorithm with varying complexity measures.}
	\begin{tabular}{p{2.3cm}|x{2.3cm}|x{2.2cm}|x{2.6cm}|x{2.2cm}	}
		Problem 		&   Variables   		 &  Tree Length 		& Visitation Length 	 &  Complexity \\ \hline
		&&&& \\ [-2ex]
		Keijzer-5		&	$0.000 \pm 0.000$	 & 	$0.004 \pm 0.003$	& 	$0.002 \pm 0.003$	 & $0.000 \pm 0.000$ \\
		Vladislavleva-1	&	$1.568 \pm 1.568$    & 	$0.047 \pm 0.062$	& 	$1.460 \pm 8.291$    & $5.509 \pm 12.00$ \\
		Vladislavleva-2	&	$0.112 \pm 0.587$	 & 	$0.023 \pm 0.022$	& 	$0.019 \pm 0.014$	 & $0.823 \pm 3.193$ \\
		Vladislavleva-7	&	$0.529 \pm 2.753$	 & 	$0.168 \pm 0.099$	& 	$0.138 \pm 0.037$	 & $0.138 \pm 0.125$ \\
		Pagie-1 		&	$0.015 \pm 2.174$	 & 	$0.445 \pm 1.737$	& 	$0.061 \pm 0.087$	 & $0.028 \pm 0.046$ \\
		Poly-10			&	$0.457 \pm 0.211$	 & 	$0.558 \pm 1.188$	& 	$0.510 \pm 0.626$	 & $0.301 \pm 0.129$ \\
		Friedman-1		&	$0.150 \pm 0.000$	 & 	$0.156 \pm 0.019$	& 	$0.169 \pm 0.024$	 & $0.175 \pm 0.021$ \\
		Friedman-2		&	$0.083 \pm 0.032$	 & 	$0.143 \pm 0.054$	& 	$0.149 \pm 0.039$	 & $0.140 \pm 0.049$ \\
		Tower			&	$0.138 \pm 0.026$	 & 	$0.144 \pm 0.020$	& 	$0.144 \pm 0.019$	 & $0.138 \pm 0.018$ \\
	\end{tabular} 
	\label{tab:results_test}
\end{table}

%% file: results_length.tex
\begin{table}
	\centering
	\caption{Average and standard deviation ($\mu \pm \sigma$) of the length of the best training individuals for 50 repetitions of NSGA-II with varying complexity measures.}
	\begin{tabular}{p{2.3cm}|x{2.3cm}|x{2.2cm}|x{2.6cm}|x{2.2cm}	}
		Problem 		&   Variables   	 &  Tree Length 	 	 & Visitation Length 	 &  Complexity \\ \hline
		&&&& \\ [-2ex]
		Keijzer-5	 	&  $86.3 \pm 19.34$	 &  $21.2 \pm 13.30$	 &  $22.7 \pm 15.75$	 &  $47.0 \pm 18.59$ \\ 
		Vladislavleva-1	&  $91.8 \pm 15.27$	 &  $44.6 \pm 22.72$	 &  $41.7 \pm 24.71$	 &  $85.7 \pm 22.22$ \\ 
		Vladislavleva-2	&  $88.7 \pm 16.81$	 &  $42.6 \pm 23.47$	 &  $37.7 \pm 21.21$	 &  $79.8 \pm 21.77$ \\ 
		Vladislavleva-7	&  $94.4 \pm 10.93$	 &  $44.0 \pm 29.41$	 &  $48.4 \pm 29.98$	 &  $81.3 \pm 24.08$ \\ 
		Pagie-1	 		&  $91.7 \pm 16.50$	 &  $51.7 \pm 28.88$	 &  $40.6 \pm 23.34$	 &  $72.4 \pm 28.01$ \\ 
		Poly-10	 		&  $96.8 \pm 7.320$	 &  $53.0 \pm 30.79$	 &  $54.8 \pm 30.06$	 &  $72.7 \pm 24.70$ \\ 
		Friedman-1		&  $90.5 \pm 12.96$	 &  $61.0 \pm 26.81$	 &  $55.1 \pm 29.74$	 &  $69.2 \pm 26.53$ \\ 
		Friedman-2	 	&  $91.4 \pm 12.00$	 &  $47.1 \pm 28.60$	 &  $40.8 \pm 25.30$	 &  $68.3 \pm 25.54$ \\ 
		Tower	 		&  $94.4 \pm 8.860$	 &  $59.8 \pm 29.30$	 &  $58.9 \pm 27.02$	 &  $78.2 \pm 21.07$ \\ 	
	\end{tabular} 
	\label{tab:results_length}
\end{table}